\documentclass[10pt,twocolumn,letterpaper]{article}

\usepackage{iccv}
\usepackage{latexsym}
\usepackage{booktabs} 
\usepackage{multirow}
\usepackage{url}
\usepackage{amsfonts}
\usepackage{bm}
\usepackage{color}
\usepackage{diagbox}
\usepackage{times}
\usepackage{epsfig}
\usepackage{graphicx}
\usepackage{setspace}
\usepackage{amsmath}
\usepackage{algorithm}
\usepackage{amssymb}
\DeclareMathOperator*{\argmax}{arg\,max}

\usepackage{dsfont}
\usepackage[noend]{algcompatible}

\usepackage[breaklinks=true,bookmarks=false]{hyperref}

\iccvfinalcopy 

\def\iccvPaperID{6903} 

\ificcvfinal\fi

\begin{document}

\title{Visual Distant Supervision for Scene Graph Generation}

\author{Yuan Yao$^{1}\thanks{\quad indicates equal contribution}$\hspace{0.4em}, Ao Zhang$^{1*}$, Xu Han$^{1}$, Mengdi Li$^{2}$, \\ Cornelius Weber$^{2}$, Zhiyuan Liu$^{1}$\thanks{\quad Corresponding author: Z.Liu (liuzy@tsinghua.edu.cn)}\hspace{0.4em}, Stefan Wermter$^{2}$,  Maosong Sun$^{1}$\\
$^{1}$Department of Computer Science and Technology\\
Institute for Artificial Intelligence, Tsinghua University, Beijing, China\\
Beijing National Research Center for Information Science and Technology, China \\
$^{2}$Knowledge Technology Group, Department of Informatics, University of Hamburg, Hamburg, Germany\\
\small{\texttt{yuan-yao18@mails.tsinghua.edu.cn, zhanga6@outlook.com}}
}


\maketitle
\ificcvfinal\thispagestyle{empty}\fi

\begin{abstract}
   Scene graph generation aims to identify objects and their relations in images, providing structured image representations that can facilitate numerous applications in computer vision. However, scene graph models usually require supervised learning on large quantities of labeled data with intensive human annotation. In this work, we propose visual distant supervision, a novel paradigm of visual relation learning, which can train scene graph models without any human-labeled data. The intuition is that by aligning commonsense knowledge bases and images, we can automatically create large-scale labeled data to provide distant supervision for visual relation learning. To alleviate the noise in distantly labeled data, we further propose a framework that iteratively estimates the probabilistic relation labels and eliminates the noisy ones. Comprehensive experimental results show that our distantly supervised model outperforms strong weakly supervised and semi-supervised baselines. By further incorporating human-labeled data in a semi-supervised fashion, our model outperforms state-of-the-art fully supervised models by a large margin (e.g., $8.3$ micro- and $7.8$ macro-recall@50 improvements for predicate classification in Visual Genome evaluation). We make the data and code for this paper publicly available at \url{https://github.com/thunlp/VisualDS}.
\end{abstract}

\section{Introduction}

\begin{figure}[t]
    \centering
    \includegraphics[width=\columnwidth]{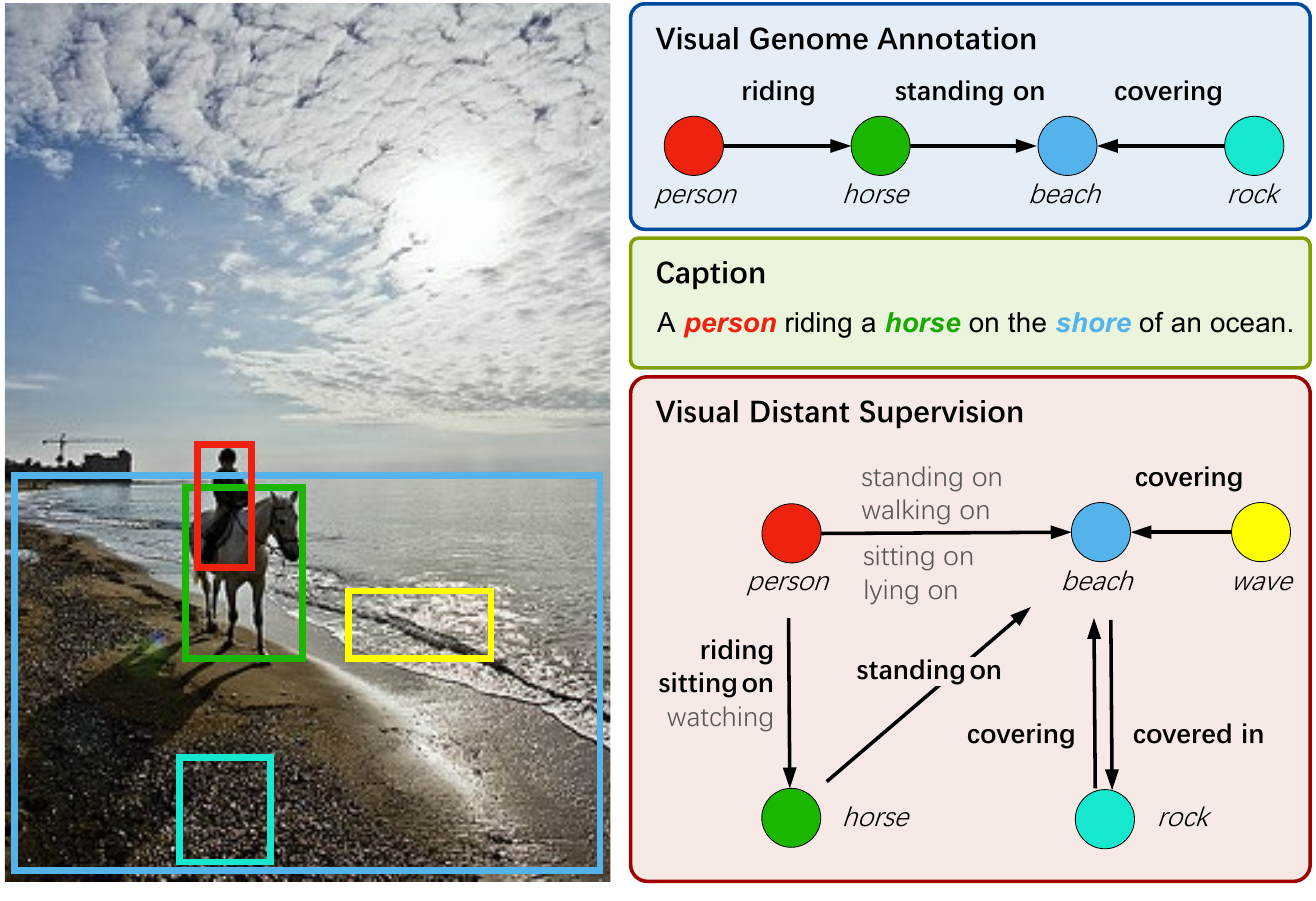}
    \caption{An example from Visual Genome~\cite{krishna2017visual} based on the refined relation schemes from Chen \etal~\cite{chen2019scene}, where human annotation from Visual Genome, weak supervision information from the corresponding caption, and raw relation labels from distant supervision are shown respectively. Correct relation labels are highlighted in bold. By aligning commonsense knowledge bases and images, visual distant supervision can create large-scale labeled data without any human efforts to facilitate visual relation learning. Best viewed in color.}
    \label{fig:example}
\end{figure}

Scene graph generation aims to identify objects and their relations in real-world images. For example, the scene graph shown in Figure~\ref{fig:example} depicts the image with several relational triples, such as (\textit{person}, \texttt{riding}, \textit{horse}) and (\textit{horse}, \texttt{standing on}, \textit{beach}). Such structured representations provide deep understanding of the semantic content of images, and have facilitated state-of-the-art models in numerous applications in computer vision, such as visual question answering~\cite{DBLP:conf/nips/HudsonM19,shi2019explainable}, image retrieval~\cite{johnson2015image,schuster2015generating}, image captioning~\cite{yang2019auto,gu2019unpaired} and image generation~\cite{johnson2018image}.



\begin{figure}[t]
    \centering
    \hspace{-2.5em}
    \includegraphics[width=0.8\columnwidth]{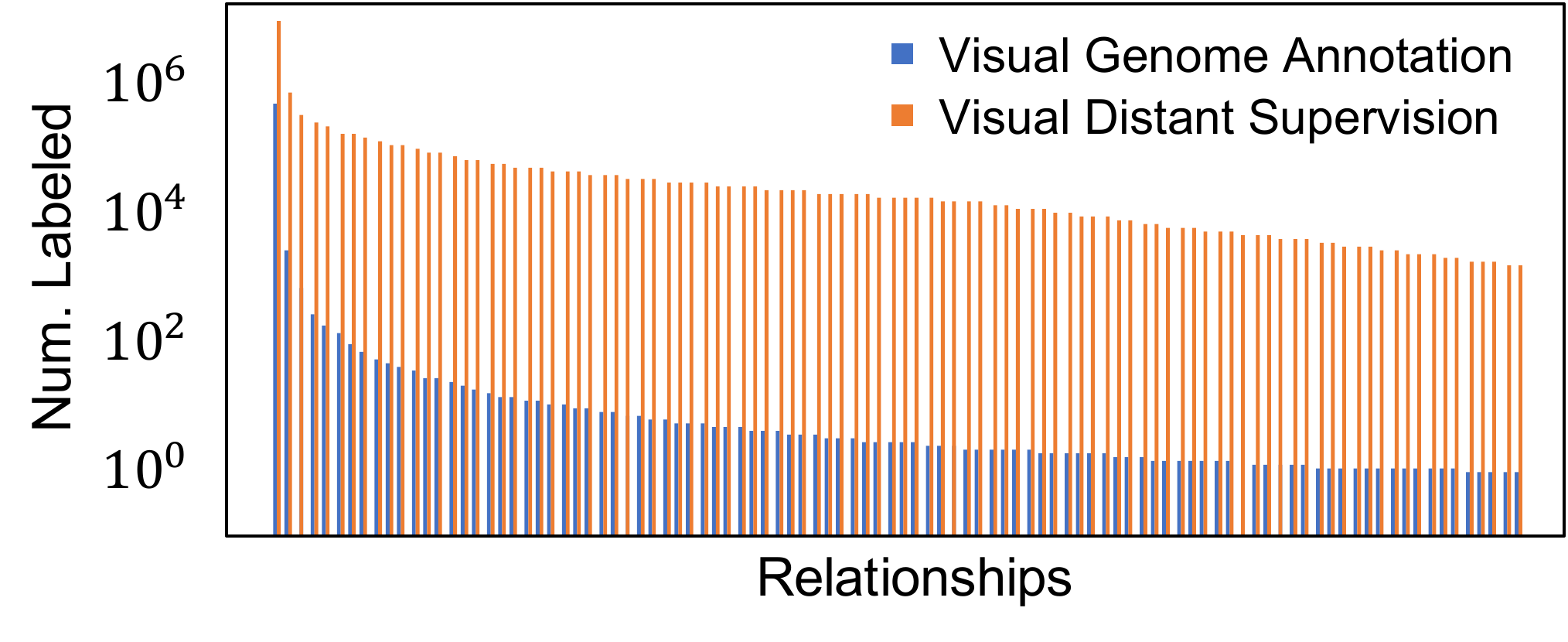}
    \caption{Number of labeled instances of top 3,000 relationships from Visual Genome annotation and visual distant supervision.}
    \label{fig:statistics}
\end{figure}

Tremendous efforts have been devoted to generating scene graphs from images~\cite{xu2017scene,li2017scene,yang2018graph,lu2016visual,zhang2017visual}. However, scene graph models usually require supervised learning on large quantities of human-labeled data. Manually constructing large-scale datasets for visual relation learning is extremely labor-intensive and time-consuming~\cite{lu2016visual,krishna2017visual}. Moreover, even with the human-labeled data, scene graph models usually suffer from the long-tail relation distribution in real-world scenarios. Figure~\ref{fig:statistics} shows the statistics on Visual Genome~\cite{krishna2017visual}, where over $98\%$ of the top $3,000$ relation categories do not have sufficient labeled instances and are thus ignored by most scene graph models. 


To address the problems, a promising direction is to utilize large-scale unlabeled data with minimal human efforts via semi-supervised or weakly supervised learning. Chen \etal~\cite{chen2019scene} propose to first learn a simple relation predictor using several human-labeled seed instances for each relation, and then assign soft labels to unlabeled data to train scene graph models. However, semi-supervised models still require human annotation that scales linearly with the number of relations. Moreover, learning from limited seed instances is vulnerable to high variance and subjective annotation bias. Some works have also explored learning from weakly supervised relation labels, which are obtained by parsing the captions of the corresponding images~\cite{zhang2017ppr,peyre2017weakly}. However, due to reporting bias~\cite{gordon2013reporting}, captions only summarize a few salient relations in images, and ignore less salient and background relations, e.g., (\textit{rock}, \texttt{covering}, \textit{beach}) in Figure~\ref{fig:example}. The resultant models will thus be biased towards a few salient relations, which cannot well serve scene graph generation that aims to exhaustively extract all reasonable relational triples in the scene.

In this work, we propose \textit{visual distant supervision}, a novel paradigm of visual relation learning, which can train scene graph models without any human-labeled data. The intuition is that commonsense knowledge bases encode relation candidates between objects, which are likely to be expressed in images. For example, as shown in Figure~\ref{fig:example}, multiple relation candidates, e.g., \texttt{riding}, \texttt{sitting on} and \texttt{watching}, can be retrieved from commonsense knowledge bases for the object pair \textit{person} and \textit{horse}, where \texttt{riding} and \texttt{sitting on} are actually expressed in the given image. By aligning commonsense knowledge bases 
and images, we can create large-scale labeled data to provide distant supervision for visual relation learning without any human efforts. Since the distant supervision is provided by knowledge bases, the relations can be exhaustively labeled between all object pairs. We note that many reasonable relation labels from distant supervision are missing in Visual Genome even after intensive human annotations, e.g., (\textit{wave}, \texttt{covering}, \textit{beach}) in Figure~\ref{fig:example}.


Moreover, distant supervision can also alleviate the long-tail problem. As shown in Figure~\ref{fig:statistics}, using the same number of images, distant supervision can produce $1$-$3$ orders of magnitude more labeled relation instances than its human-labeled counterpart. Note that the number of distantly labeled relation instances can be arbitrarily large, given the nearly unlimited image data on the Web. 


Distant supervision is convenient in training scene graph models without human-labeled data. When human-annotated data is available, distantly labeled data can also be incorporated in a semi-supervised fashion to surpass fully supervised models. We show that after pre-training on distantly labeled data, simple fine-tuning on human-labeled data can lead to significant improvements over strong fully supervised models.

Despite its potential, we note distant supervision may introduce noise in relation labels, e.g., (\textit{person}, \texttt{watching}, \textit{horse}) in Figure~\ref{fig:example}. The reason is that distant supervision only provides relation candidates based on object categories, whereas the actual relations between two objects in an image usually depend on the image content. In principle, the noise in distantly labeled data can be alleviated by maximizing the coherence between distant labels and visual patterns of object pairs. Previous works have shown that without specially designed denoising methods, neural models are capable of detecting noisy labels to some extent, and learning meaningful signals from noisy data~\cite{jiang2018mentornet,feng2018reinforcement}. In this work, to better alleviate the noise in distantly labeled data, we further propose a framework that iteratively estimates the probabilistic relation labels and eliminates noisy ones. The framework can be realized by optimizing the coherence of internal statistics of distantly labeled data, and can also be seamlessly integrated with external semantic signals (e.g., image-caption retrieval models), or human-labeled data to achieve better denoising results.

Comprehensive experimental results show that, without using any human-labeled data, our distantly supervised model outperforms strong weakly supervised and semi-supervised baseline methods. By further incorporating human-labeled data in a semi-supervised fashion, our model outperforms state-of-the-art fully supervised models by a large margin (e.g., $8.3$ micro- and $7.8$ macro-recall@50 improvements for predicate classification task in Visual Genome evaluation). Based on the experiments, we discuss multiple promising directions for future research.

Our contributions are threefold: (1) We propose visual distant supervision, a novel paradigm of visual relation learning, which can train scene graph models without any human-labeled data, and also improve fully supervised models. (2) We propose a denoising framework to alleviate the noise in distantly labeled data. (3) We conduct comprehensive experiments which demonstrate the effectiveness of visual distant supervision and the denoising framework.

\section{Related Work}

\smallskip
\noindent
\textbf{Visual Relation Detection.} Identifying visual relations between objects is critical for image understanding, which have received broad attention from the community~\cite{lu2016visual,zhang2017visual,gupta2008beyond,gupta2009observing,desai2012detecting,ramanathan2015learning,chao2015hico,yatskar2016situation}. Johnson \etal~\cite{johnson2015image} further formulate scene graphs that encode all objects and their relations in images into structured graph representations. Tremendous efforts have been devoted to generating scene graphs, including refining contextualized graph features~\cite{dai2017detecting,xu2017scene,li2017scene,zellers2018neural}, developing computationally efficient scene graph models~\cite{li2018factorizable,yang2018graph,zhang2017ppr} and designing effective loss functions~\cite{yang2018graph,zhang2019graphical}. However, scene graph models usually require supervised learning on large amounts of human-labeled data~\cite{lu2016visual,krishna2017visual}.

\smallskip
\noindent
\textbf{Weakly Supervised Scene Graph Generation.}  To alleviate the heavy reliance on human-labeled data, recent works on scene graph generation have explored semi-supervised and weakly supervised learning methods. Chen \etal~\cite{chen2019scene} propose to bootstrap scene graph models from several human-labeled seed instances for each relation, which still requires manual labor and is vulnerable to high variance. Other works attempt to obtain weakly supervised relation labels from the corresponding image captions. Peyre \etal~\cite{peyre2017weakly} propose to ground the labels to object pairs by imposing global grounding constraints~\cite{foulds2010review}. To improve the computational efficiency, Zhang \etal~\cite{zhang2017ppr} design a network branch to select a pair of object proposals for each relation label. Baldassarre \etal~\cite{DBLP:conf/eccv/BaldassarreSSA20} propose to first detect the relation via graph networks, and then recover the subject and object of the predicted relation. Zareian \etal~\cite{zareian2020weakly} reformulate scene graphs as bipartite graphs of objects and relations, and align the predicted graphs to their weakly supervised labels. However, since the weakly supervised relation labels are parsed from the corresponding captions, the resultant models will be biased towards the most salient relations, ignoring many less salient and background relations.

\begin{figure*}[t]
    \centering
    \definecolor{olive}{RGB}{108,134,22}
    \definecolor{red}{RGB}{159,12,16}
    \includegraphics[width=0.86\textwidth]{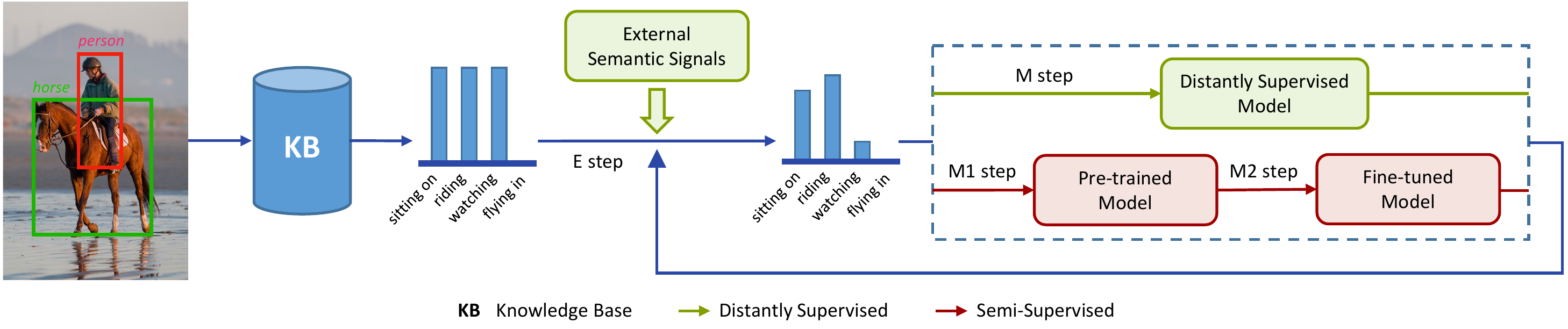}
    \vspace{0.5em}
    \caption{The denoising framework for visual distant supervision. The framework iteratively estimates the probabilistic relation labels based on EM optimization, and can be realized in distantly supervised and semi-supervised fashion. Best viewed in color.}
    \label{fig:framework}
\end{figure*}

\smallskip
\noindent
\textbf{Textual Distant Supervision.} In natural language processing, there has been a long history of extracting relational triples from text (i.e., textual relation extraction) to complete knowledge bases~\cite{huffman1995learning,mooney1999relational,zeng2014relation,zhang2015bidirectional}. Supervised textual relation extraction models are usually limited by the sizes of human-annotated datasets. To address the issue, Mintz \etal~\cite{mintz2009distant} propose to align Freebase~\cite{bollacker2008freebase}, a world knowledge base, to text to provide distant supervision for textual relation extraction. Although both targeting at extracting relations, we provide distant supervision for visual relation learning by aligning commonsense knowledge bases with visual concepts, in contrast to textual distant supervision that aligns world knowledge bases with textual entities. 

\smallskip
\noindent
\textbf{Learning with Noisy Labels.} Visual distant supervision may introduce noisy relation labels, which may hurt the performance of scene graph models. In textual distant supervision, many denoising methods have been developed under the multi-instance learning formulation~\cite{zeng2015distant,lin2016neural,zhang2017position,han2018hierarchical}. However, visual relation detection aims to extract relations on instance level (i.e., predicting relation instances in specific images), whereas textual relation extraction focuses on extracting global relations between entities (i.e., synthesizing information from all sentences containing the entity pair to identify their global relation). Therefore, denoising methods for distantly supervised textual relation extraction cannot well serve visual relation detection. Previous instance-level denoising methods have explored handling noisy labels in image classification~\cite{jiang2018mentornet,ren2018learning,sugiyama2018co,thulasidasan2019combating,li2019learning} and object detection~\cite{li2020learning,shen2020noise,gao2019note} based on internal data statistics. In comparison, our denoising framework cannot only leverage internal data statistics, but can also be seamlessly integrated with external semantic signals and human-labeled data for better denoising results.

\section{Problem Definition}
We first provide a formal definition of the problem and key terminologies in our work.

\smallskip
\noindent
\textbf{Scene Graphs.} Formally, a scene graph consists of the following elements: (1) Objects. Each object $obj = (b, c)$ is associated with a bounding box $b \in \mathbb{R}^4$ and a category $c \in \mathcal{C}$, where $\mathcal{C}$ is the object category set. (2) Relations, with $r \in \mathcal{R}$, where $\mathcal{R}$ is the relation category set (including \texttt{NA} indicating no relation). Given an image, scene graph models aim to extract the relational triple $(s, r, o)$. 

\smallskip
\noindent
\textbf{Knowledge Bases.} Most knowledge bases store relations between concepts in the form of relational triples $(c_i, r, c_j)$. 


\smallskip
\noindent
\textbf{Distantly Supervised and Semi-supervised Learning.} In the traditional fully supervised relation learning, human labeled data $D_L$ is required to train scene graph models. In distantly supervised relation learning, where no human-labeled data is available, we automatically create distantly labeled data $D_S$ using images and knowledge bases to train scene graph models. When human-labeled data $D_L$ is available, we can further leverage $D_S \cup D_L$ in a semi-supervised fashion to surpass fully supervised models trained on $D_L$.

\section{Visual Distant Supervision}
\label{sec:visual_distant_supervision}
In this section, we introduce the assumption and approach of visual distant supervision, which aims to create large-scale labeled data for visual relation learning. 

The key insight of visual distant supervision is that visual relational triples correspond to commonsense knowledge. For example, the relational triple (\textit{person}, \texttt{riding}, \textit{horse}) expresses the commonsense ``\textit{person can ride horses}''. Therefore, commonsense knowledge bases can provide possible relation candidates between visual objects to distantly supervise visual relation learning. To this end, we perform visual distant supervision by first constructing a commonsense knowledge base, and then aligning it to images.

\smallskip
\noindent
\textbf{Knowledge Base Construction.} Although several commonsense knowledge bases have been constructed~\cite{speer2017conceptnet}, we find that they cannot well serve visual distant supervision due to their incompleteness. Therefore, instead of adopting existing knowledge bases, we automatically construct a commonsense knowledge base by extracting relational triples from Web-scale image captions. Specifically, we extract relational triples from Conceptual Captions~\cite{sharma2018conceptual}, which contains $3.3$M captions of images, using a rule-based textual parser~\cite{schuster-etal-2015-generating}. The resultant knowledge base contains $18,618$ object categories, $63,232$ relation categories and $1,876,659$ distinct relational triples, where each object pair has $1.94$ relations on average.

\smallskip
\noindent
\textbf{Knowledge Base and Image Alignment.} To align the knowledge base and images, we need to obtain the bounding boxes and categories of objects in each image. In this work, we utilize the images and object annotations from Visual Genome, while obtaining object information in open-domain images using object detectors~\cite{ren2015faster} is also applicable. After that, for each object pair, we retrieve all the relation labels in the knowledge base as relation candidates. 

Nevertheless, we observe that directly performing distant supervision will produce considerable noise. For example, if there are multiple \textit{person} and \textit{horse} objects in an image, there will be a \texttt{riding} relation label between each \textit{person} and \textit{horse} pair. Inspired by previous works~\cite{zellers2018neural}, we adopt a simple but effective heuristic constraint to filter out a large number of noisy labels. Specifically, we assign distant relation labels for an object pair only if the bounding boxes of the subject and object have overlapping areas. After alignment, the relation labels from distant supervision can cover $70.3\%$ relation labels from Visual Genome.


\section{Denoising for Visual Distant Supervision}

The relation labels from distant supervision can be readily used to train any scene graph models. However, distant supervision may introduce noisy relation labels, which may hurt the model performance. To alleviate the noise in visual distant supervision, we propose a denoising framework, as shown in Figure~\ref{fig:framework}. Regarding the ground-truth relation labels of distantly labeled data as latent variables, we iteratively estimate the probabilistic relation labels, and eliminate the noisy ones to train any scene graph models. The framework can be realized by optimizing the coherence of internal statistics of distantly labeled data, and can also be seamlessly integrated with external semantic signals (e.g., image-caption retrieval models), or human-labeled data to achieve better denoising results. In this section, we introduce the framework in distantly supervised and semi-supervised settings respectively. We refer readers to the appendix for the pseudo-code of the framework.

\subsection{Distantly Supervised Framework}
\label{sec:distantly_supervised_framework}

In distantly supervised framework, where only distantly labeled data $D_S$ is available, we aim to refine the probabilistic relation labels of $D_S$ iteratively by maximizing the coherence of its internal statistics using EM optimization. 

\smallskip
\noindent
\textbf{E step.} In the E step of $t$-th iteration, we estimate the labels of distantly labeled data to obtain $D_S^t = \{(s, \mathbf{r}^t, o)^{(k)}\}_{k=1}^N$, where $\mathbf{r}^t$ indicates the latent relation between the object pair $(s, o)$ in an image.\footnote{We omit the superscript $k$ in the following for simplicity.} Specifically, $\mathbf{r}^t \in \mathbb{R}^{|\mathcal{R}|}$ is a probabilistic distribution over all relations in $\mathcal{R}$, which comes from either (1) raw labels from distant supervision (in the initial iteration), or (2) probabilistic relation labels estimated by the model. Given an object pair $(s,o)$, we denote the set of retrieved distant labels as $\mathcal{R}_{(s,o)}$. Note that during the EM optimization, we only refine the distant labels in $\mathcal{R}_{(s,o)}$, and keep $\mathbf{r}_i^t=0$, if $r_i \not\in \mathcal{R}_{(s,o)}$.

\smallskip
(1) In the \textbf{initial iteration} (i.e., $t=1$), the relation labels are assigned by aligning knowledge bases and images (see Section~\ref{sec:visual_distant_supervision}), denoted as follows:

\begin{equation}
\label{eq:raw_distant_labels}
\mathbf{d} = \Psi(s, o, \Lambda),
\end{equation}
where $\Lambda$ is the knowledge base, $\Psi(\cdot)$ is the alignment operation. $\mathbf{d}$ is a multi-hot vector where $\mathbf{d}_i=1$ if $r_i \in \mathcal{R}_{(s,o)}$ and otherwise $0$. 

We argue that when available, external semantic signals are useful in distinguishing reasonable distant labels from noisy ones. Without losing generality, in this work, we adopt CLIP~\cite{radford2learning}, a state-of-the-art cross-modal representation model pre-trained on large-scale image-caption pairs\footnote{In the distantly supervised framework, we are careful to not introduce human annotated relation data in any component. The knowledge base is constructed using a rule-based method from image captions, and CLIP is pre-trained on image-caption pairs only.}, to measure the semantic relatedness between a textual relational triple from distant supervision, and the corresponding visual object pair. Specifically, given an object pair, we obtain the visual input by masking the area in the image that is not covered by the bounding boxes of the object pair. To obtain the textual input, we simply concatenate the subject, relation and object in the relational triple into a text snippet. Then the visual and textual inputs are fed into CLIP to obtain their unnormalized relatedness score (i.e., cosine similarity), summarized as follows:

\begin{equation}
\alpha_i = \Phi(v, u_i),
\end{equation}
where $v$ is the visual input of the object pair, $u_i$ is the textual input of the distantly labeled relation $r_i$, $\Phi(\cdot, \cdot)$ denotes external semantic signals, and $\alpha_i$ is the relatedness score. After that, we normalize the relatedness score to obtain the probabilistic relation distribution over $\mathcal{R}_{(s,o)}$:


\begin{equation}
\label{eq:normalize}
\mathbf{e}_i = \frac{\exp(\alpha_i)}{\sum_{j=1}^{|\mathcal{R}|} \mathds{1}[\mathbf{d}_j=1]\exp(\alpha_j)}, \ \ r_i \in \mathcal{R}_{(s,o)},
\end{equation}
where $\mathds{1}[x]$ is $1$ if $x$ is true otherwise $0$. $\mathbf{e}$ is the probabilistic relation distribution given by external semantic signals, where $\mathbf{e}_i=0$ if $r_i \not\in \mathcal{R}_{(s,o)}$. 

The relation distribution can then be initialized by $\mathbf{r}^1 = \mathbf{e}$. Note that external signals are not necessarily required by the framework (i.e., initialize $\mathbf{r}^1 = \mathbf{d}$ when such external signals are not available). 

\smallskip
(2) In the \textbf{non-initial iterations} (i.e., $t>1$), we infer the probabilistic relation distribution by the convex combination of the internal prediction from scene graph models, and external semantic signals:


\begin{equation}
\mathbf{r}_i^{t} = \omega f_i(s, o; \theta^{t-1}) + (1-\omega)\mathbf{e}_i,
\end{equation}
where $f_i(s, o; \theta^{t-1})$ is the probability of $r_i$ from the scene graph model with parameter $\theta^{t-1}$. Here $f_i(s, o; \theta^{t-1})$ is obtained by normalizing the relation logits over $\mathcal{R}_{(s,o)}$ as in Equation~\ref{eq:normalize}. $\omega \in [0,1]$ is a weighting hyperparameter, where $\omega=1$ when external signals are not available.

We note it is possible that none of the distant labels between $(s, o)$ are correct (see (\textit{person}, \textit{beach}) in Figure~\ref{fig:example} for example). Therefore, we eliminate noisy object pairs from $D_S^t$, by discarding object pairs with top $k\%$ \texttt{NA} relation logits given by the scene graph model.

\smallskip
\noindent
\textbf{M step.} In the M step, given the distant labels from the E step, we optimize the scene graph model parameters $\theta^{t-1}$ by maximizing the log-likelihood of $D_S^t$ as follows:


\begin{equation}
    \theta^{t} = \argmax _\theta \mathcal{L}_p(D_S^t; \theta^{t-1}), 
\end{equation}
where $\mathcal{L}_p(D_S^t; \theta^{t-1})$ is the entropy-based log-likelihood function, which incorporates the probabilistic relation distribution of $D_S^t$ in a noise-aware approach as follows:


\begin{equation}
\nonumber
\small
\begin{split}
    \mathcal{L}_p(D_S^t; \theta^{t-1}) = & \sum_{(s, \mathbf{r}^t, o) \in D_S^t} \sum_{i=1}^{|\mathcal{R}|} \mathbf{r}_i^t (\mathds{1}[\mathbf{d}_i=1] \log f_i(s, o; \theta^{t-1}) \\
    &+ \mathds{1}[\mathbf{d}_i = 0] \log (1- f_i(s, o; \theta^{t-1}))), \hspace{2.35em} \normalsize{\text{(6)}} \vspace{0.5em}   
\end{split}
\end{equation}
\setcounter{equation}{6}
where $\theta^0$ is randomly initialized.

\begin{table*}
    \begin{center}
    \resizebox{\linewidth}{!}{%
    \small
    \begin{tabular}{ll cccc cccc cccc c}
    \toprule
    & \hspace{-0.9em} \multirow{2}{*}{Models} & \multicolumn{4}{c}{Predicate Classification}& \multicolumn{4}{c}{Scene Graph Classification} & \multicolumn{4}{c}{Scene Graph Detection} & \multirow{2}{*}{Mean}\\
    \cmidrule(lr){3-6} \cmidrule(lr){7-10} \cmidrule(lr){11-14}
    &  & R@50 & R@100 & mR@50 & mR@100 & R@50 & R@100 & mR@50 & mR@100 & R@50 & R@100 & mR@50 & mR@100 & \\
    \midrule
    
    \hspace{-0.8em} \parbox[t]{2mm}{\multirow{6}{*}{\rotatebox[origin=c]{90}{Baselines}}}
 & \hspace{-0.9em} Freq~\cite{zellers2018neural}* & 20.80 & 20.98 & - & - & 10.92 & 11.08 & - & - & 11.01 & 11.64 & - & - & - \\
 & \hspace{-0.9em}    Freq-Overlap~\cite{zellers2018neural}* & 20.90  & 22.21 & - & - & \hspace{1.6mm}9.91 & \hspace{1.6mm}9.91 & - & - & 10.84 & 10.86 & - & - & -\\
 &  \hspace{-0.9em}   Decision Tree~\cite{quinlan1986induction}* &  33.02  &  33.35  &  -  &  -  &  14.51  &  14.57  &  -  &  -  &  12.58  &  13.23  &  -  &  -  & - \\
 &  \hspace{-0.9em}   Label Propagation~\cite{zhu2002learning}* &  25.17  &  25.41  &  -  &  -  &  \hspace{1.6mm}9.91  &  \hspace{1.6mm}9.97  &  -  &  -  &  \hspace{1.6mm}6.74  &  \hspace{1.6mm}6.83  &  -  &  -  & -\\
  &  \hspace{-0.9em}   Weak Supervision\dag & 44.96 & 47.19 & 24.58 & 27.14 & 19.27 & 19.93 & \hspace{1.6mm}6.97 & \hspace{1.6mm}7.54 & 19.78 & 21.33 & \hspace{1.6mm}5.01 & \hspace{1.6mm}5.41 & 20.76 \\
 &  \hspace{-0.9em}   Limited Labels~\cite{chen2019scene} &  49.68  &  50.73  &  37.43  &  38.91  &  24.65  &  25.08  &  13.30  &  13.94  &  22.87  &  24.16  &  12.66  &  13.39 & 27.23 \\
    \midrule
    
    
    \hspace{-0.8em} \parbox[t]{2mm}{\multirow{6}{*}{\rotatebox[origin=c]{90}{DS (Ours)}}}
 & \hspace{-0.9em}    EXT & \hspace{1.6mm}6.64 & \hspace{1.6mm}9.74 & 10.66 & 15.16 & \hspace{1.6mm}3.96 & \hspace{1.6mm}4.82 & \hspace{1.6mm}4.25 & \hspace{1.6mm}4.92 & \hspace{1.6mm}1.93 & \hspace{1.6mm}3.06 & \hspace{1.6mm}1.66 & \hspace{1.6mm}2.49 & \hspace{1.6mm}5.77 \\
 & \hspace{-0.9em}    Raw Label & 30.61 & 33.48 & 20.98 & 23.25 & 15.69 & 16.99 & 11.06 & 12.53 & \hspace{1.6mm}9.36 & 10.26 & \hspace{1.6mm}6.56 & \hspace{1.6mm}7.13 & 16.49 \\
 & \hspace{-0.9em}    Raw Label + EXT  & 38.21 & 40.90 & 24.94 & 27.45 & 17.52 & 18.85 & 11.66 & 12.56 & 15.84 & 18.31 & \hspace{1.6mm}9.49 & 11.23 & 20.58 \\
 & \hspace{-0.9em}    Motif\dag & 48.88 & 51.73 & 34.40 & 39.69 & 23.15 & 24.18 & 15.81 & 16.66 & 18.73 & 22.10 & 10.89 & 13.34 & 26.63 \\
  & \hspace{-0.9em}   Motif & 50.23 & 53.18 & 33.99 & 40.62 & 24.90 & 26.00 & 16.50 & 18.03 & 20.09 & 22.74 & 12.21 & 14.42 & 27.74 \\
 & \hspace{-0.9em}    Motif + DNS + EXT&  \textbf{53.40} & \textbf{56.54} & \textbf{37.68} & \textbf{41.98} & \textbf{26.12} & \textbf{27.46} & \textbf{17.20} & \textbf{18.39} & \textbf{23.69} & \textbf{25.59} & \textbf{13.84} & \textbf{15.23} & \textbf{29.76} \\
    \midrule
    \hspace{-0.8em} \parbox[t]{2mm}{\multirow{1}{*}{\rotatebox[origin=c]{90}{FS}}}
 & \hspace{-0.9em}   
 Motif~\cite{zellers2018neural} & 67.93 & 70.20 & 52.65 & 55.41 & 31.14 & 31.92 & 23.53 & 25.27 & 28.90 & 31.25 & 18.26 & 20.63 & 38.09 \\
 \midrule
 \hspace{-0.8em} \parbox[t]{2mm}{\multirow{2}{*}{\rotatebox[origin=c]{90}{SS}}}
 & \hspace{-0.9em}  Motif + Pretrain (Ours)  & 73.22 & 75.04 & \textbf{60.44} & \textbf{63.67} & 34.11 & 34.88 & 26.51 & 27.94 & 30.70 & 33.32 & \textbf{24.76} & 27.45 & 42.67 \\

 & \hspace{-0.9em}  Motif + DNS (Ours) & \textbf{76.28} & \textbf{77.98} & 60.20 & 63.61 & \textbf{35.93} & \textbf{36.47} & \textbf{28.07} & \textbf{30.09} & \textbf{33.94} & \textbf{37.26} & 23.90 & \textbf{28.06} & \textbf{44.31} \\
    \bottomrule
    \end{tabular}
    }
    \end{center}
    \caption{Main results of visual distant supervision (\%). DS: distantly supervised, FS: fully supervised, SS: semi-supervised. EXT: external semantic signal, DNS: denoising. * denotes results from Chen \etal~\cite{chen2019scene}, \dag~indicates models trained on the images with captions.}
    \label{table:main results}
\end{table*}

\subsection{Semi-supervised Framework}

Distantly supervised models can be further integrated with human-labeled data to surpass fully supervised models. In fact, we find after pre-training on distantly supervised data (see Section~\ref{sec:distantly_supervised_framework}), simple fine-tuning on human-labeled data can lead to significant improvements over fully supervised models. This simple pre-training and fine-tuning paradigm is appealing, since it does not change the \textit{number of parameters}, \textit{architectures} and \textit{overhead} in training specific downstream scene graph models.

Nevertheless, we find closely integrating human-labeled data in the denoising framework can yield better performance, since coherence can be achieved between distantly labeled data $D_S$ and human-labeled data $D_L$ for mutual enhancement. Our semi-supervised framework largely follows the distantly supervised framework in Section~\ref{sec:distantly_supervised_framework}, where we estimate probabilistic relation labels in E step, and optimizing model parameters in M step. To integrate human-labeled data, we further decompose the M step into two sub-steps: pre-training on distantly labeled data (M1 step), and fine-tuning on human-labeled data (M2 step).

\smallskip
\noindent
\textbf{E step.} In the E step of $t$-th iteration, we estimate the labels of distantly supervised data $D_S^t$. Here $\mathbf{r}_t$ is obtained by (1) first obtaining raw distant labels $\mathbf{d}$ as in Equation~\ref{eq:raw_distant_labels}, and (2) then estimating probabilistic relation labels by the fine-tuned scene graph model as follows:


\begin{equation}
\label{eq:E_step_semi_supervised}
\mathbf{r}_i^{t} = f_i(s, o; \theta_2^{t-1}), \ \ r_i \in \mathcal{R}_{(s,o)},
\end{equation}
where $f_i(\cdot; \theta_2^{t-1})$ is the fine-tuned scene graph model from the M2 step of the previous iteration. In the initial iteration, $f_i(\cdot; \theta_2^{0})$ is initialized by a fully supervised model. Note Equation~\ref{eq:E_step_semi_supervised} does not include external semantic signals, since models fine-tuned on human-labeled data can provide more direct denoising signals. After that, we discard noisy object pairs (see Section~\ref{sec:distantly_supervised_framework}). To better cope with the fine-tuning procedure on human-labeled data, where models are usually optimized towards a single discrete relation label between an object pair, we discretize $\mathbf{r}^{t}$ into a one-hot vector $\hat{\mathbf{r}}^{t}$, where $\hat{\mathbf{r}}_i^{t}=1$, if $i=\argmax_j \mathbf{r}_j^t$. 

\smallskip
\noindent
\textbf{M1 step.} After obtaining the distant label $\hat{\mathbf{r}}^{t}$ from the E step, we pre-train the scene graph model from scratch: $\theta^{t}_1 = \argmax _\theta \mathcal{L}_q(D_S^t; \theta)$, where $\mathcal{L}_q$ is the cross-entropy based objective as follows:


\begin{equation}
    \mathcal{L}_q(D_S^t; \theta) = \sum_{(s, \hat{\mathbf{r}}^t, o) \in D_S^t} \sum_{i=1}^{|\mathcal{R}|} \mathds{1}[\hat{\mathbf{r}}_i^t=1] \log f_i(s, o; \theta).
\end{equation}

\smallskip
\noindent
\textbf{M2 step.} In the M2 step, we simply fine-tune the pre-trained scene graph model on the human labeled data with $\theta^{t}_2 = \argmax _\theta \mathcal{L}_q(D_L; \theta^t_1)$.

\section{Experiments}
In this section, we empirically evaluate visual distant supervision and the denoising framework on scene graph generation. We also show the advantage of visual distant supervision in dealing with long-tail problems, and its promising potential when equipped with ideal knowledge bases.

\subsection{Experimental Settings}
We first introduce the experimental settings, including datasets, evaluation metrics and baselines. 

\smallskip
\noindent
\textbf{Datasets.} We evaluate our models on Visual Genome~\cite{krishna2017visual}, a widely adopted benchmark for scene graph generation~\cite{xu2017scene,zellers2018neural,chen2019scene,zareian2020weakly}. Each image in the dataset is manually annotated with objects (bounding boxes and object categories) and relations. In our experiments, during training distant supervision is performed using the intersection of relations from Visual Genome and the knowledge base. During evaluation, in the main experiments, we adopt the refined relation schemes and data split from Chen \etal~\cite{chen2019scene}, which removes hypernyms and redundant synonyms in the most frequent $50$ relation categories in Visual Genome, resulting in $20$ well-defined relation categories. We also report experimental results on the Visual Genome dataset with $50$ relation categories in appendix. We refer readers to the appendix for more details about data statistics.


\smallskip
\noindent
\textbf{Evaluation Metrics.} Following previous works~\cite{xu2017scene,zellers2018neural,chen2019scene}, we assess our approach in three standard evaluation modes: (1) Predicate classification. Given the ground-truth bounding boxes and categories of objects in an image, models need to predict the predicates (i.e., relations) between object pairs. (2) Scene graph classification. Given the ground-truth bounding boxes of objects, models need to predict object categories and relations. (3) Scene graph detection. Given an image, models are asked to predict bounding boxes and categories of objects, and relations between objects. We adopt the widely used micro-recall@K (\textbf{R@K}) metric to evaluate the model performance~\cite{xu2017scene,zellers2018neural,chen2019scene}, which calculates the recall in top K relation predictions. To investigate the model performance in dealing with long-tail relations, we also report macro-recall@K (\textbf{mR@K})~\cite{chen2019knowledge,tang2019learning}, which calculates the mean recall of all relations in top K predictions. Following Zellers \etal~\cite{zellers2018neural}, we also report the mean of these metrics to show the overall performance.

\begin{table*}
    \begin{center}
    \resizebox{\linewidth}{!}{%
    \small
    \begin{tabular}{ll cccc cccc cccc c}
    \toprule
    & \hspace{-0.9em} \multirow{2}{*}{Models} & \multicolumn{4}{c}{Predicate Classification}& \multicolumn{4}{c}{Scene Graph Classification} & \multicolumn{4}{c}{Scene Graph Detection} & \multirow{2}{*}{Mean}\\
    \cmidrule(lr){3-6} \cmidrule(lr){7-10} \cmidrule(lr){11-14}
    &  & R@50 & R@100 & mR@50 & mR@100 & R@50 & R@100 & mR@50 & mR@100 & R@50 & R@100 & mR@50 & mR@100 \\
    \midrule

    \hspace{-0.8em} \parbox[t]{2mm}{\multirow{7}{*}{\rotatebox[origin=c]{90}{DS}}}
 & \hspace{-0.9em} \footnotesize{Motif} & 50.23 & 53.18 & 33.99 & 40.62 & 24.90 & 26.00 & 16.50 & 18.03 & 20.09 & 22.74 & 12.21 & 14.42 & 27.74 \\
 & \hspace{-0.9em}    \footnotesize{Motif + Cleanness Loss~\cite{li2020learning}} & 51.10 & 54.23 & 34.69 & 42.67 & 24.06 & 24.98 & 16.46 & 18.56 & 21.94 & 23.89 & 13.21 & 14.49 & 28.36 \\
 &  \hspace{-0.9em}   \footnotesize{Motif + DNS} \scriptsize{(iter 1)} & 50.23 & 53.18 & 33.99 & 40.62 & 24.90 & 26.00 & 16.50 & 18.03 & 20.09 & 22.74 & 12.21 & 14.42 & 27.74 \\
 &  \hspace{-0.9em}  \hspace{2.0em} \footnotesize{+ DNS} \scriptsize{(iter 2)} & 51.54 & 54.53 & 36.93 & 41.97 & 24.81 & 26.08 & 16.13 & 17.56 & 22.83 & 24.36 & 13.48 & 14.45 & 28.72 \\
 
 &  \hspace{-0.9em}   \footnotesize{Motif + DNS + EXT} \scriptsize{(iter 1)} & 52.82 & 55.98 & 36.25 & 41.66 & 25.79 & 26.98 & \textbf{17.39} & \textbf{18.56} & 22.63 & 25.12 & 13.30 & \textbf{15.40} & 29.32 \\
 &  \hspace{-0.9em}  \hspace{2.0em} \footnotesize{+ DNS + EXT} \scriptsize{(iter 2)} & \textbf{53.40} & \textbf{56.54} & \textbf{37.68} & \textbf{41.98} & \textbf{26.12} & \textbf{27.46} & 17.20 & 18.39 & \textbf{23.69} & \textbf{25.59} & \textbf{13.84} & 15.23 & \textbf{29.76} \\
\midrule

\hspace{-0.8em} \parbox[t]{2mm}{\multirow{1}{*}{\rotatebox[origin=c]{90}{FS}}}
 & \hspace{-0.9em} \footnotesize{Motif~\cite{zellers2018neural}} & 67.93 & 70.20 & 52.65 & 55.41 & 31.14 & 31.92 & 23.53 & 25.27 & 28.90 & 31.25 & 18.26 & 20.63 & 38.09   \\
 \midrule
 \hspace{-0.8em} \parbox[t]{2mm}{\multirow{2}{*}{\rotatebox[origin=c]{90}{SS}}}
 & \hspace{-0.9em} \footnotesize{Motif + DNS \scriptsize{(iter 1)}} & 73.50 & 75.33 & \textbf{61.40} & \textbf{65.20} & 35.39 & 35.98 & \textbf{28.71} & \textbf{30.25} & \textbf{34.83} & \textbf{37.68} & \textbf{24.78} & 27.90 & 44.25 \\
 & \hspace{-0.9em} \hspace{2.0em} \footnotesize{+ DNS \scriptsize{(iter 2)}} & \textbf{76.28} & \textbf{77.98} & 60.20 & 63.61 & \textbf{35.93} & \textbf{36.47} & 28.07 & 30.09 & 33.94 & 37.26 & 23.90 & \textbf{28.06} & \textbf{44.31} \\

    \bottomrule
    \end{tabular}
    }
    \end{center}
    \caption{Experimental results of denoising visual distant supervision (\%). Results of different denoising iterations are shown. DS: distantly supervised, FS: fully supervised, SS: semi-supervised. EXT: external semantic signal, DNS: denoising.}
    \vspace{-0.5em}
    \label{table:denoising results}
\end{table*}

\smallskip
\noindent
\textbf{Baselines.} We compare our models with strong baselines. (1) The first series of baselines learn visual relations from a few (i.e., $10$) human-labeled seed instances for each relation. Frequency-based baseline (\textbf{Freq})~\cite{zellers2018neural} predicts the most frequent relation between an object pair. Overlap-enhanced frequency-based baseline (\textbf{Freq-Overlap})~\cite{zellers2018neural} further filters out non-overlapping object pairs. Following Chen \etal~\cite{chen2019scene}, we also compare with learning \textbf{Decision Tree}~\cite{quinlan1986induction} from seed instances. (2) For \textit{semi-supervised methods} that further incorporate unlabeled data, following Chen \etal~\cite{chen2019scene}, we compare with \textbf{Label Propagation}~\cite{zhu2002learning} that propagates the labels of the seed data to unlabeled data based on data point communities. \textbf{Limited Labels}~\cite{chen2019scene} is the state-of-the-art semi-supervised scene graph model, which first learns a relational generative model using seed instances, and then assigns soft labels to unlabeled data to train scene graph models. (3) We also compare with strong \textit{weakly supervised models} (\textbf{Weak Supervision}\dag) that are supervised by the relation labels parsed from the captions of the corresponding images~\cite{peyre2017weakly,zhang2017ppr}. Specifically, we use images with captions in Visual Genome to train the weakly supervised model. We label the object pairs with relations parsed from the corresponding caption, and employ the overlapping constraint to filter out noisy labels (see Section~\ref{sec:visual_distant_supervision}). For fair comparisons, we also train a distantly supervised model without denoising based on the same images with captions in Visual Genome (\textbf{Motif}\dag). (4) For \textit{fully supervised methods}, we compare with the strong and widely adopted Neural Motif (\textbf{Motif})~\cite{zellers2018neural}. For fair comparisons, all the neural models in our experiments are implemented based on the Neural Motif model, with ResNeXt-101-FPN~\cite{lin2017feature,xie2017aggregated} as the backbone. (5) For \textit{denoising baselines}, we adapt \textbf{Cleanness Loss}~\cite{li2020learning} that heuristically down-weight the relation labels with large loss. We refer readers to the appendix for more implementation details.

\smallskip
\noindent
\textbf{Ablations.} To investigate the contribution of each component, we conduct ablation study. (1) In \textit{distantly supervised setting}, we perform distant supervision based on the general knowledge base from Section~\ref{sec:visual_distant_supervision}. \textbf{Raw Label} predicts relations by raw distant relation labels. \textbf{EXT} indicates external semantic signals. \textbf{Motif} denotes training on raw distant relation labels, and \textbf{DNS} denotes denoising based on the proposed framework. (2) In \textit{semi-supervised setting}, in addition to distantly labeled data, we assume access to full human-annotated relation data. \textbf{Pretrain} indicates directly fine-tuning the model pre-trained on distantly supervised data from the general knowledge base. \textbf{DNS} denotes denoising based on the targeted knowledge base constructed from Visual Genome training annotations. 


\subsection{Effect of Visual Distant Supervision}

We report the main results of visual distant supervision in Table~\ref{table:main results}, from which we have the following observations: (1) Without using any human-labeled data, our denoised distantly supervised model significantly outperforms all baseline methods, including weakly supervised methods and even strong semi-supervised approaches that utilize human-labeled seed data. (2) By further incorporating human-labeled data, our semi-supervised models consistently outperform state-of-the-art fully supervised models by a large margin, e.g., $8.3$ R@50 improvement for predicate classification. Specifically, simple fine-tuning of the pre-trained model can lead to significant improvement. Since the model is pre-trained on general knowledge bases, it can also be directly fine-tuned on any other scene graph dataset to achieve strong performance. Moreover, by closely integrating human-labeled data and distantly labeled data in the denoising framework, we can achieve even better performance. (3) Notably, our models achieve competitive macro-recall, which shows that our models are not biased towards a few frequent relations, and can better deal with the long-tail problem. In summary, visual distant supervision can effectively create large-scale labeled data to facilitate visual relation learning in both distantly supervised and semi-supervised scenarios.

\subsection{Effect of the Denoising Framework}

The experimental results of denoising distant supervision are shown in Table~\ref{table:denoising results}, from which we observe that: Equipped with the proposed denoising framework, our models show consistent improvements over baseline models in both distantly supervised and semi-supervised settings. Specifically, in distantly supervised setting, the model performance improves with the iterative optimization of the coherence of internal data statistics. Further incorporating external semantic signals and human-labeled data cannot only boost the denoising performance, but also speed up the convergence of the iterative algorithm. The reason is that external semantic signals and human-labeled data can provide strong auxiliary denoising signals for both better initialization and iteration of the framework. The results show that the proposed denoising framework can effectively alleviate the noise in visual distant supervision in both distantly supervised and semi-supervised settings.

\subsection{Analysis}

\noindent
\textbf{Distant Supervision with Ideal Knowledge Bases.} The effectiveness of visual distant supervision may be limited by the incompleteness of knowledge bases, and mismatch in relation and object names between knowledge bases and images. We show the potential of visual distant supervision with ideal knowledge bases. Specifically, we construct an ideal knowledge base from the training annotations of Visual Genome, which better covers the relational knowledge in the dataset, and can be well aligned to Visual Genome images. Experimental results are shown in Table~\ref{table:ideal_KB}, from which we observe that: Equipped with ideal knowledge bases, the performance of distantly supervised models improves significantly. Notably, the macro-recall of the denoised distantly supervised model dramatically improves, e.g., with $13.2$ absolute gain in mR@50, \textit{achieving comparable macro performance to fully supervised approaches}. Therefore, we expect visual distant supervision will even better promote visual relation learning, given that knowledge bases are becoming increasingly complete.

\begin{table}
    \begin{center}
    \resizebox{\linewidth}{!}{%
    \small
    \begin{tabular}{ll|cccc}
    \toprule
    & \hspace{-0.9em} Models & R@50 & R@100 & mR@50 & mR@100 \\
    \midrule
    
    \hspace{-0.8em} \parbox[t]{2mm}{\multirow{4}{*}{\rotatebox[origin=c]{90}{\footnotesize{DS (Ours)}}}}
 & \hspace{-0.9em} \footnotesize{Raw Label} & 35.62\scriptsize{ (+5.01)} & 39.78\scriptsize{ (+6.30)} & 34.83\scriptsize{ (+13.85)} & 39.45\scriptsize{ (+16.20)} \\
 &  \hspace{-0.9em}  \footnotesize{Raw Label + EXT}  & 45.07\scriptsize{ (+6.86)} & 49.00\scriptsize{ (+8.10)} & 44.18\scriptsize{ (+19.24)} & 48.56\scriptsize{ (+21.11)} \\
 & \hspace{-0.9em}    \footnotesize{Motif} & 53.02\scriptsize{ (+2.79)} & 56.31\scriptsize{ (+3.13)} & 46.65\scriptsize{ (+12.66)} & 50.90\scriptsize{ (+10.28)} \\
 & \hspace{-0.9em}    \footnotesize{Motif + DNS + EXT} & \textbf{55.54}\scriptsize{ (+2.14)} & \textbf{58.99}\scriptsize{ (+2.45)} & \textbf{50.87}\scriptsize{ (+13.19)} & \textbf{55.69}\scriptsize{ (+13.71)} \\

\midrule

\hspace{-0.8em} \parbox[t]{2mm}{\multirow{1}{*}{\rotatebox[origin=c]{90}{\footnotesize{FS}}}}
 & \hspace{-0.9em} \footnotesize{Motif~\cite{zellers2018neural}} & \hspace{-7.6mm}67.93 & \hspace{-7.7mm}70.02 & \hspace{-9.0mm}52.65 & \hspace{-9.0mm}55.41  \\
    
    \bottomrule
    \end{tabular}
    }
    \end{center}
    \caption{Experimental results of distant supervision with ideal knowledge bases on predicate classification (\%). We also show the absolute improvements with respect to the results from general knowledge bases. DS: distantly supervised, FS: fully supervised.}
    \vspace{-0.5em}
    \label{table:ideal_KB}
\end{table}

\smallskip
\noindent
\textbf{Human Evaluation.} Since relations in existing scene graph datasets are typically not exhaustively annotated, previous works have concentrated on evaluating models with recall metric~\cite{xu2017scene,zellers2018neural,chen2019scene}. To provide multi-dimensional evaluation, we further evaluate the precision of our scene graph models. Specifically, we randomly sample $200$ images from the validation set of Visual Genome. For each image, we obtain $20$ top-ranking relational triples extracted by a scene graph model. Then we ask human annotators to label whether each relational triple is correctly identified from the image. We show the human evaluation results in Table~\ref{table:human_evaluation}, where micro- and macro-precision@K are reported. (1)~For raw labels from distant supervision, in addition to the $10.9\%$ strictly correct labels in top $20$ recommendations,  we find that $32.6\%$ of the distant labels are plausible, which also provides useful signals for visual relation learning. (2)~The denoised distantly supervised model achieves competitive precision. Moreover, compared to recall evaluation, our semi-supervised model exhibits more significant relative advantage over the fully supervised model. The results show that with appropriate denoising, distant supervision can lead scene graph models to strong performance in both precision and recall.


\begin{table}
    \begin{center}
    \resizebox{0.81\linewidth}{!}{%
    \small
    \begin{tabular}{ll|cccc}
    \toprule
    & \hspace{-0.9em} Models & P@10 & P@20 & mP@10 & mP@20 \\
    \midrule
    
    \hspace{-0.8em} \parbox[t]{2mm}{\multirow{2}{*}{\rotatebox[origin=c]{90}{DS}}}
 & \hspace{-0.9em} Raw Labels & 12.07 & 10.85 & 11.41 & 12.04 \\
 &  \hspace{-0.9em} Motif + DNS + EXT & \textbf{31.93} & \textbf{25.29} & \textbf{24.79} & \textbf{20.79} \\
 \midrule
 \hspace{-0.8em} \parbox[t]{2mm}{\multirow{1}{*}{\rotatebox[origin=c]{90}{FS}}} & \hspace{-0.9em}  Motif~\cite{zellers2018neural} & 42.22 & 31.09 & 39.60 & 29.22 \\
 \midrule
 \hspace{-0.8em}
 \parbox[t]{2mm}{\multirow{1}{*}{\rotatebox[origin=c]{90}{SS}}} & \hspace{-0.9em}  Motif + DNS & \textbf{50.58} & \textbf{38.68} & \textbf{47.49} & \textbf{38.52} \\
    
    \bottomrule
    \end{tabular}
    }
    \end{center}
    \caption{Human evaluation results on predicate classification (\%), where micro- and macro-precision@K are reported. DS: distantly supervised, FS: fully supervised, SS: semi-supervised.}
    \vspace{-0.5em}
    \label{table:human_evaluation}
\end{table}

\section{Discussion and Outlook}
In this work, we show the promising potential of visual distant supervision in visual relation learning. In the future, given the recent advances in textual relation extraction~\cite{lin2016neural,han2018fewrel,yao2019docred}, we believe the following research directions are worth exploring: (1) Developing more sophisticated denoising methods to better realize the potential of visual distant supervision. (2) Completing commonsense knowledge bases by extracting \textit{global} relations between object pairs from multiple images. (3) Reducing the overhead and bias in annotating large-scale visual relation learning datasets based on raw labels from distant supervision.


\section{Conclusion}

In this work, we propose visual distant supervision and a denoising framework for visual relation learning. Comprehensive experiments demonstrate the effectiveness of visual distant supervision and the denoising framework. In this work, we denoise distant labels for each object pair in isolation. In the future, we will explore modeling the holistic coherence of the produced scene graph to better alleviate the noise in visual distant supervision. It is also important to investigate whether visual distant supervision will introduce extra bias to scene graph models.

\section{Acknowledgement}
This work is jointly funded by the Natural Science Foundation of China (NSFC) and the German Research Foundation (DFG) in Project Crossmodal Learning, NSFC 62061136001 / DFG TRR-169. Yao is also supported by 2020 Tencent Rhino-Bird Elite Training Program.


{\small
\bibliographystyle{ieee_fullname}
\bibliography{iccv}
}

\clearpage

\appendix

\section{The Denoising Framework: Pseudo-Code}
In this section, we provide the pseudo-code of the denoising framework in distantly supervised and semi-supervised settings respectively.

\begin{table}[h]
\small
    \centering
    \renewcommand\arraystretch{1.0}
    \begin{tabular}{l}
    \toprule
    \hspace{-0.5em} \textbf{Algorithm 1} Distantly Supervised Denoising Framework  \\
    \midrule
    \textbf{Require:} $\Lambda$: commonsense knowledge base   \\
    \textbf{Require:} $D_S$: distantly labeled image data \\
    \textbf{Require:} $f(\cdot; \theta)$: any scene graph model, with parameter $\theta$ \\
    \textbf{Optional:} $\Phi$: external semantic signal 
    \\
    \hspace{0.4em} 1: Randomly initialize the model parameter  $\theta^{0}$ \\
    \hspace{0.4em} 2: // Initial E step: estimate the probabilistic distant labels \\
    \hspace{0.4em} 3: Obtain distant labels $\mathbf{d} = \Psi(s, o, \Lambda)$\\
    \hspace{0.4em} 4: \textbf{if} external signal $\Phi$ available \textbf{then}\\
    \hspace{0.4em} 5: \hspace{1em}Initialize $\mathbf{r}^1=\mathbf{e}$ \\ 
    \hspace{0.4em} 6: \textbf{else} \\
    \hspace{0.4em} 7: \hspace{1em}Initialize $\mathbf{r}^1=\mathbf{d}$ \\ 
    \hspace{0.4em} 8: \textbf{end if} \\
    \hspace{0.4em} 9: // Initial M step: optimize model parameter \\
    \hspace{0.2em}10: Optimize $\theta^{1} = \argmax _\theta \mathcal{L}(D_S^1; \theta^{0})$ \\
    
    \hspace{0.2em}11: \textbf{while} not done \textbf{do} \\
    \hspace{0.2em}12: \hspace{1em}// E step: estimate the probabilistic distant labels \\
    \hspace{0.2em}13: \hspace{1em}\textbf{if} external signal $\Phi$ available \textbf{then}\\
    \hspace{0.2em}14: \hspace{1em} \hspace{1em} Estimate $\mathbf{r}_i^{t} = \omega f_i(s, o; \theta^{t-1}) + (1-\omega)\mathbf{e}_i$ \\ 
    \hspace{0.2em}15: \hspace{1em}\textbf{else} \\
    \hspace{0.2em}16: \hspace{1em} \hspace{1em} Estimate $\mathbf{r}_i^{t} = f_i(s, o; \theta^{t-1})$\\
    \hspace{0.2em}17: \hspace{1em}\textbf{end if} \\
    \hspace{0.2em}18: \hspace{1em}Eliminate noisy object pairs\\
    \hspace{0.2em}19: \hspace{1em}// M step: optimize model parameter \\
    \hspace{0.2em}20: \hspace{1em}Optimize $\theta^{t} = \argmax _\theta \mathcal{L}_p(D_S^t; \theta^{t-1})$ \\
    
    \hspace{0.2em}21: \textbf{end while} \\
    
    \bottomrule
    \end{tabular}
    \label{table:distantly supervised algorithm}
\end{table}

\begin{table}[t]
\small
    \centering
    \renewcommand\arraystretch{1.0}
    \begin{tabular}{l}
    \toprule
    \hspace{-0.5em} \textbf{Algorithm 2} Semi-Supervised Denoising Framework  \\
    \midrule
    \textbf{Require:} $\Lambda$: commonsense knowledge base   \\
    \textbf{Require:} $D_S$: distantly labeled image data \\
    \textbf{Require:} $D_L$: human-labeled image data \\
    \textbf{Require:} $f(\cdot; \theta)$: any scene graph model, with parameter $\theta$ \\
    \hspace{0.4em} 1: Initialize $f(\cdot; \theta^0)$ with fully supervised model   \\
    \hspace{2.5em} $\theta^{0} = \argmax _\theta \mathcal{L}_q(D_L; \theta)$ \\
    
    \hspace{0.4em} 2: \textbf{while} not done \textbf{do} \\
    \hspace{0.4em} 3: \hspace{1em}// E step: estimate the probabilistic distant labels \\
    \hspace{0.4em} 4: \hspace{1em}Estimate $\mathbf{r}_i^{t} = f_i(s, o; \theta_2^{t-1})$\\
    \hspace{0.4em} 5: \hspace{1em}Eliminate noisy object pairs\\
    \hspace{0.4em} 6: \hspace{1em}// M1 step: pre-train on distantly labeled data $D_S^t$ \\
    \hspace{0.4em} 7: \hspace{1em}Optimize $\theta^{t}_1 = \argmax _\theta \mathcal{L}_q(D_S^t; \theta)$ \\
    \hspace{0.4em} 8: \hspace{1em}// M2 step: fine-tune on human-labeled data $D_L$ \\
    \hspace{0.4em} 9: \hspace{1em}Optimize $\theta^{t}_2 = \argmax _\theta \mathcal{L}_q(D_L; \theta^t_1)$ \\

    \hspace{0.2em}10: \textbf{end while} \\
    
    \bottomrule
    \end{tabular}
    \label{table:semi-supervised algorithm}
\end{table}

\begin{figure*}[t]
    \centering
    \includegraphics[width=1.0\textwidth]{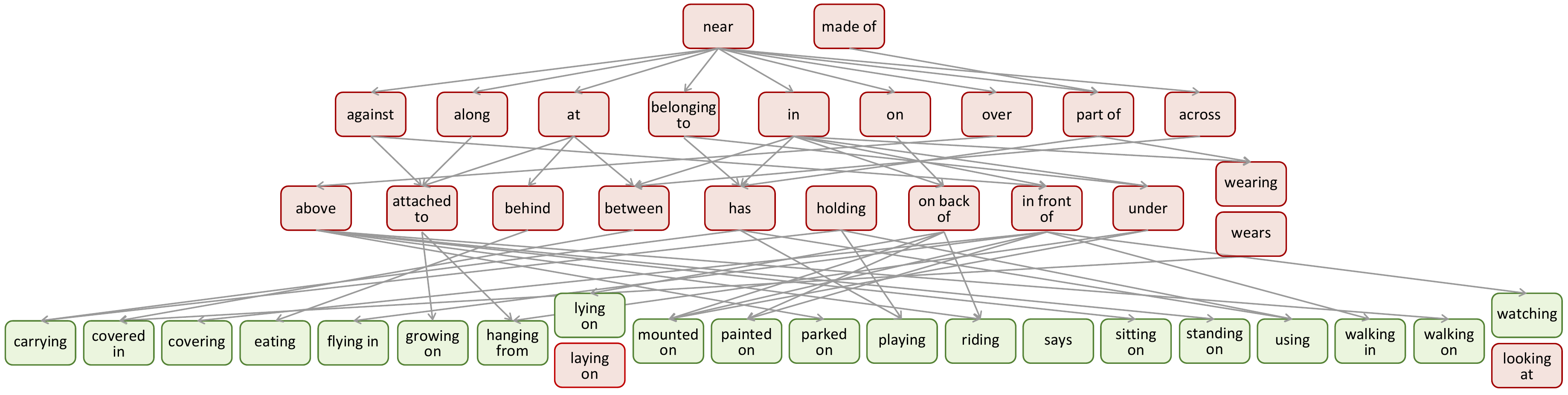}
    \caption{Dependencies of Visual Genome relations from Chen \etal~\cite{chen2019scene}. Directed arrows: hypernyms. Stacked nodes: synonyms. Red nodes: removed relations. Green nodes: retained relations.}
    \label{fig:relation schemes}
    \vspace{0.5em}
\end{figure*}

\begin{table*}
    \begin{center}
    \resizebox{\linewidth}{!}{%
    \small
    \begin{tabular}{ll cccc cccc cccc c}
    \toprule
    & \hspace{-0.9em} \multirow{2}{*}{Models} & \multicolumn{4}{c}{Predicate Classification}& \multicolumn{4}{c}{Scene Graph Classification} & \multicolumn{4}{c}{Scene Graph Detection} & \multirow{2}{*}{Mean}\\
    \cmidrule(lr){3-6} \cmidrule(lr){7-10} \cmidrule(lr){11-14}
    &  & R@50 & R@100 & mR@50 & mR@100 & R@50 & R@100 & mR@50 & mR@100 & R@50 & R@100 & mR@50 & mR@100 & \\
    \midrule
    \hspace{-0.8em} \parbox[t]{2mm}{\multirow{4}{*}{\rotatebox[origin=c]{90}{DS (Ours)}}}
 & \hspace{-0.9em}    Raw Label & 16.93 & 19.02 & \hspace{1.6mm}5.75 & \hspace{1.6mm}7.15 & 11.62 & 12.59 & 4.01 & 4.64 & \hspace{1.6mm}7.52 & \hspace{1.6mm}7.79 & 2.32 & 2.49 & \hspace{1.6mm}8.49 \\
  & \hspace{-0.9em}    Motif  & 33.21 & 36.17 & \textbf{10.84} & \textbf{12.48} & 20.35 & 21.85 & \textbf{5.23} & \textbf{5.91} & 12.89 & 15.48 & \textbf{4.51} & \textbf{5.58} & 15.38 \\
 & \hspace{-0.9em}    Motif + DNS  & 35.53 & 38.28 & \hspace{1.6mm}9.35 & 10.74 & 21.33 & 22.70 & 4.79 & 5.36 & 15.04 & 17.86 & 3.83 & 4.66 & 15.79 \\
 & \hspace{-0.9em}    Motif + DNS + EXT & \textbf{36.43} & \textbf{39.21} & \hspace{1.6mm}8.68 & 10.03 & \textbf{21.88} & \textbf{23.21} & 3.80 & 4.23 & \textbf{16.32} & \textbf{18.78} & 3.82 & 4.55 & \textbf{15.91} \\
    \midrule
    \hspace{-0.8em} \parbox[t]{2mm}{\multirow{1}{*}{\rotatebox[origin=c]{90}{FS}}}
 & \hspace{-0.9em}    Motif~\cite{zellers2018neural} & 63.96 & 65.93 & 15.15 & 16.24 & 38.04 & 38.90 & 8.66 & 9.25 & \textbf{31.00} & 35.06 & 6.66 & 7.73 & 28.05 \\
 \midrule
 \hspace{-0.8em}
 \parbox[t]{2mm}{\multirow{1}{*}{\rotatebox[origin=c]{90}{SS}}}
 & \hspace{-0.9em}  Motif + DNS (Ours) & \textbf{64.43} & \textbf{66.43} & \textbf{16.12} & \textbf{17.47} & \textbf{38.38} & \textbf{39.25} & \textbf{9.27} & \textbf{9.86} & 30.91 & \textbf{35.08} & \textbf{7.03} & \textbf{8.29} & \textbf{28.54} \\
    \bottomrule
    \end{tabular}
    }
    \end{center}
    \caption{Results of visual distant supervision on Visual Genome 50 predicates (\%). DS: distantly supervised, SS: semi-supervised. }
    \label{table:50 predicate results}
    \vspace{0.5em}
\end{table*}

\begin{table*}[!h]
    \begin{center}
    \resizebox{0.75\linewidth}{!}{%
    \small
    \begin{tabular}{ll ccc ccc ccc}
    \toprule
    & \hspace{-0.9em} \multirow{2}{*}{Models} & \multicolumn{3}{c}{Accuracy}& \multicolumn{3}{c}{Mean Accuracy} & \multicolumn{3}{c}{\# Non-zero Predicates}\\
    \cmidrule(lr){3-5} \cmidrule(lr){6-8} \cmidrule(lr){9-11}
    & & top-1 & top-5 & top-10 & top-1 & top-5 & top-10 & top-1 & top-5 & top-10 \\
    \midrule
    
    \hspace{-0.8em} \parbox[t]{2mm}{\multirow{1}{*}{\rotatebox[origin=c]{90}{FS}}}
 & \hspace{-0.9em} Motif~\cite{zellers2018neural} & 64.05 & 81.35 & 85.05 & 1.51 & 5.15 & \hspace{-0.5em}7.32 & 103 & 169 & 212  \\
 \midrule
 \hspace{-0.8em}
 \parbox[t]{2mm}{\multirow{1}{*}{\rotatebox[origin=c]{90}{SS}}} & \hspace{-0.9em}    Motif + DNS (Ours) & \textbf{66.10} & \textbf{84.26} & \textbf{87.72} & \textbf{3.08} & \textbf{9.49} & \textbf{13.44} & \textbf{218} & \textbf{362} & \textbf{435} \\
    \bottomrule
    \end{tabular}
    }
    \end{center}
    \caption{Results of visual distant supervision on Visual Genome 1,700 predicates (\%). FS: fully supervised, SS: semi-supervised.}
    \label{table:1700 predicate results}
\end{table*}

\section{Implementation Details}
In this section, we provide implementation details of our model and baseline methods. For fair comparisons, all the neural models in our experiments are implemented using the same object detector, scene graph model and backbone.

\smallskip
\noindent
\textbf{Object Detector.} We adopt the object detector implementation from Tang \etal~\cite{tang2020unbiased}. Specifically, the object detector is trained using SGD optimizer with learning rate $8 \times 10^{-3}$ and batch size $8$. During the training process, the learning rate is decreased two times by $10$ in $30,000$ and $40,000$ iterations respectively.

\smallskip
\noindent
\textbf{Scene Graph Model.} For the base scene graph model, we follow the implementation of Neural Motif~\cite{zellers2018neural}, with ResNeXt-101-FPN~\cite{lin2017feature,xie2017aggregated} as the backbone.
For predicate classification and scene graph classification, the ratio of positive relation samples and negative relation samples in each image is at most $1$:$3$. For scene graph detection, a relational triplet is considered as positive only if the detected object pairs match ground-truth annotations, i.e., with identical object categories and bounding box IoU $> 0.5$. We find that this strict constraint leads to sparse positive supervision in experiments, especially in our distantly supervised setting. To address the issue, we change the ratio of positive and negative relation samples in distantly supervised setting to strictly $1$:$1$. During evaluation, we only keep 64 object bounding box predictions. The models are trained using SGD optimizer on $2$ NVIDIA GeForce RTX 2080 Ti, with momentum $0.9$, batch size $12$ and weight decay $5 \times 10^{-4}$. 

\smallskip
\noindent
\textbf{Our Model.} 
All the hyperparameters of our model are selected by grid search on the validation set. (1) In \textit{distantly supervised setting}, for the denoising framework, the weighting hyperparameter $\omega$ is $0.9$, and $75\%$ noisy object pairs are discarded. In the first iteration, we train the model with learning rate $0.12$, and decrease the learning rate $3$ times after the plateaus of validation performance. In the second iteration, the learning rate is $0.012$, and decays $1$ times after the validation performance plateaus. Note that following Devlin \etal~\cite{devlin-etal-2019-bert}, the learning rate in the second iteration is smaller than the first iteration, since we are actually fine-tuning the model parameter inherited from the first iteration. (2) In \textit{semi-supervised setting}, for the denoising framework, no object pairs are discarded. The initial fully supervised model is trained with learning rate $0.12$. In both iterations, the learning rate is $0.12$ for pre-training, and $0.012$ for fine-tuning. The learning rate decays $2$ and $1$ times in the first and second iterations respectively. To fine-tune the pre-trained distantly supervised model without semi-supervised denoising, we optimize with learning rate $0.012$ that decays $2$ times. The decay rate is $10$ for all models.

\smallskip
\noindent
\textbf{Baselines.} For the Limited Labels~\cite{chen2019scene}, we train the decision trees for $200$ different trails on $10$ randomly sampled human-labeled seed instances for each relation, and select the best models according to the performance on the validation set. For the weakly supervised model, since Visual Genome does not have image-level captions, we utilize all the images in Visual Genome training set that have captions from COCO~\cite{lin2014microsoft}, resulting in $35,340$ images with captions in total. Then we train the weakly supervised model with all Visual Genome object annotations from these images, and relation labels parsed from the corresponding captions. For the Cleanness Loss~\cite{li2020learning}, we denoise with soft weight given by the confidence of the scene graph model.

\section{Data statistics}
 In our main experiments, we adopt the refined relation schemes from Chen \etal~\cite{chen2019scene}, which removes hypernyms (e.g., \texttt{near} and \texttt{on}), redundant synonyms (e.g., \texttt{lying on} and \texttt{laying on}), and unclear relations (e.g., \texttt{and}) in the most frequent $50$ relation categories in Visual Genome, resulting in $20$ well-defined relation categories. The relation dependencies from Chen \etal~\cite{chen2019scene} are shown in Figure~\ref{fig:relation schemes}. The dataset contains $10,986$, $1,566$ and $3,025$ images in training, validation and test set respectively, where each image contains an average of $13.58$ objects, $2.10$ human-labeled relation instances and $15.60$ distantly labeled relation instances.

\section{Supplementary Experiments}
\begin{figure*}[t]
    \centering
    \definecolor{olive}{RGB}{108,134,22}
    \definecolor{red}{RGB}{159,12,16}
    \includegraphics[width=1.0\textwidth]{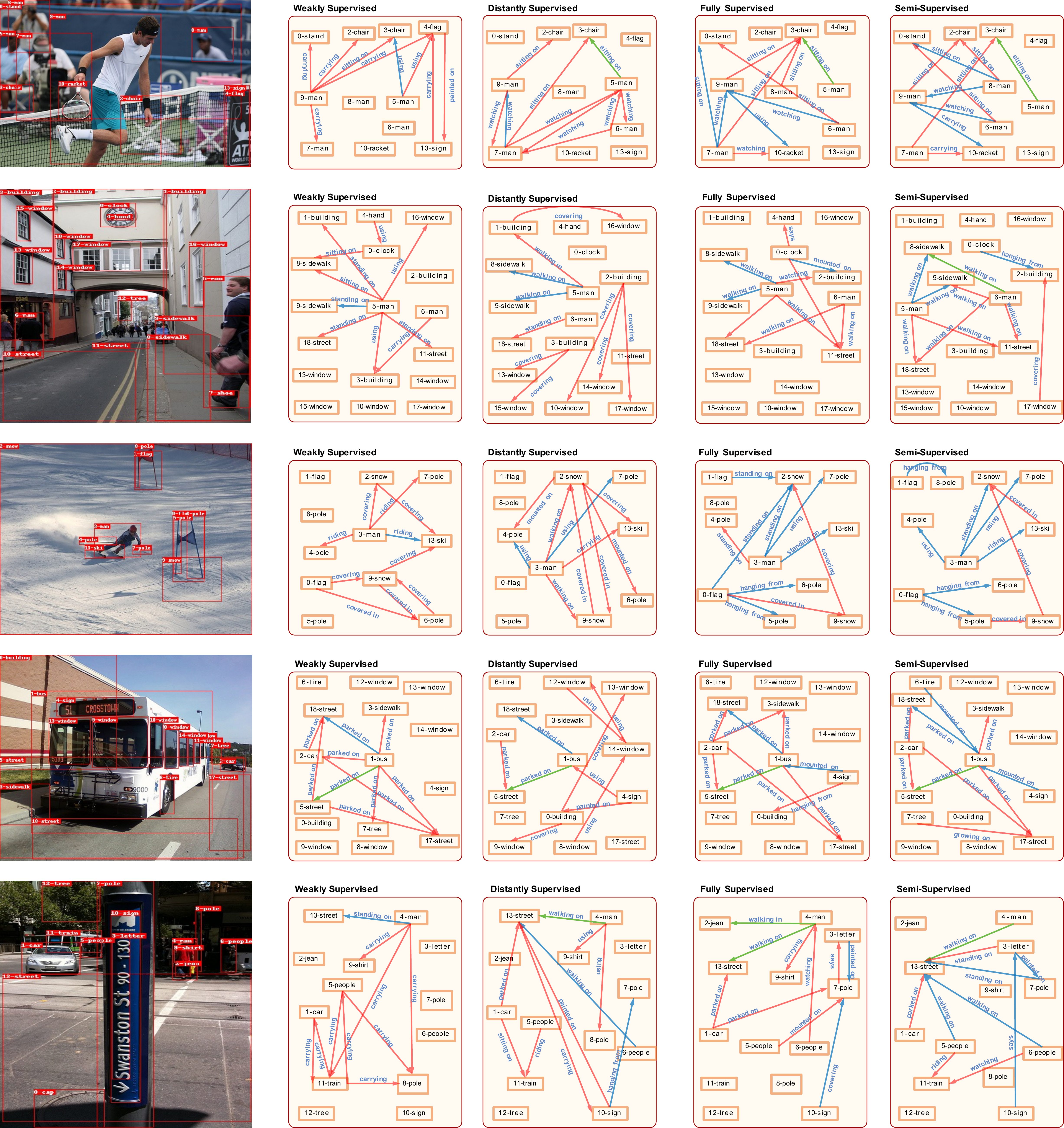}
    \vspace{-0.8em}
    \caption{Qualitative examples of model predictions in predicate classification task. We show top 10 predictions from (1) models that do not utilize human-labeled data, including weakly supervised and distantly supervised model, and (2) models that leverage human-labeled data, including fully supervised model and our semi-supervised model. Green edges: predictions that match Visual Genome annotations, blue edges: plausible predictions that are not labeled in Visual Genome, red edges: implausible predictions.}
    \label{fig:case appendix}
\end{figure*}

\smallskip
\noindent
\textbf{Case Study.} We provide qualitative examples in Figure~\ref{fig:case appendix} for better understanding of different scene graph models.

\smallskip
\noindent
\textbf{Results on 50 Visual Genome Predicates.} We report the experimental results on 50 Visual Genome predicates in Table~\ref{table:50 predicate results}. We observe that although reasonable performance can be achieved, the improvement from distant supervision and the denoising framework shrinks. This is because that the $50$ relations are not well-defined, where the major relations are problematic hypernym (e.g., \texttt{near} and \texttt{on}), redundant synonym (e.g., \texttt{lying on} and \texttt{laying on}), and unclear (e.g. \texttt{and}) relations, as pointed out by Chen \etal~\cite{chen2019scene}. The problematic relation schemes can bring difficulties to denoising distant supervision.

\smallskip
\noindent
\textbf{Results on 1,700 Visual Genome Predicates.} Visual distant supervision can alleviate the long-tail problem and therefore enables large-scale visual relation extraction. To investigate the effectiveness of visual distant supervision in handling large-scale visual relations, we refine the full Visual Genome predicates following the principles proposed by~\cite{chen2019scene}, resulting in $1,700$ well-defined predicates. In addition to top-K accuracy, to better focus on the long-tail performance, we also report top-K mean accuracy and the number of non-zero predicates (i.e., predicates with at least one correctly predicted instance). From the experimental results in Table~\ref{table:1700 predicate results}, we observe that our model significantly outperforms its fully supervised counterpart. Notably, our model nearly doubles the top-K mean accuracy and the number of non-zero predicates, demonstrating the promising potential of visual distant supervision in handling large-scale visual relations in the future.

\end{document}